\documentclass[10pt,twocolumn,letterpaper]{article}

\usepackage{iccv}
\usepackage{times}
\usepackage{epsfig}
\usepackage{graphicx}
\usepackage{amsmath}
\usepackage{amssymb}

\usepackage{cite}
\usepackage{amsfonts}
\usepackage{algorithmicx}
\usepackage{algpseudocode}
\usepackage[linesnumbered,ruled,vlined]{algorithm2e}
\usepackage{textcomp}
\usepackage{xcolor}
\usepackage{lipsum}
\usepackage{breqn}
\usepackage{verbatim}

\usepackage{caption}
\usepackage{subcaption}
\usepackage[accsupp]{axessibility}  

\usepackage[breaklinks=true,bookmarks=false]{hyperref}

\iccvfinalcopy 

\def\iccvPaperID{4} 

\ificcvfinal\fi

\begin{document}

\title{FOX-NAS: Fast, On-device and Explainable Neural Architecture Search}

\author{
Chia-Hsiang Liu$^1$,
Yu-Shin Han$^1$,
Yuan-Yao Sung$^1$,
Yi Lee$^1$,
Hung-Yueh Chiang$^2$,
Kai-Chiang Wu$^1$, 
\\
$^1$National Yang Ming Chiao Tung University, 
$^2$The University of Texas at Austin
\\
{\tt\small \{jacoblau.cs08g, yushinhan.eic09g\}@nctu.edu.tw, hungyueh.chiang@utexas.edu,}~\\{\tt\small kcw@cs.nctu.edu.tw}
}

\maketitle
\ificcvfinal\thispagestyle{empty}\fi

\begin{abstract}
   Neural architecture search can discover neural networks with good performance, and One-Shot approaches are prevalent. One-Shot approaches typically require a supernet with weight sharing and predictors that predict the performance of architecture. However, the previous methods take much time to generate performance predictors thus are inefficient. To this end, we propose FOX-NAS that consists of fast and explainable predictors based on simulated annealing and multivariate regression. Our method is quantization-friendly and can be efficiently deployed to the edge. The experiments on different hardware show that FOX-NAS models outperform some other popular neural network architectures. For example, FOX-NAS matches MobileNetV2 and EfficientNet-Lite0 accuracy with 240\% and 40\% less latency on the edge CPU. Search code and pre-trained models are released at  \url{https://github.com/great8nctu/FOX-NAS}. ~\footnote{FOX-NAS is the 3rd place winner of the 2020 Low-Power Computer Vision Challenge (LPCVC), DSP classification track. See all evaluation results at \url{https://lpcv.ai/competitions/2020}.}
\end{abstract}

\section{Introduction}
Deep learning has been applied in various fields in the past decade, including image classification~\cite{szegedy2015going,simonyan2014very}, object detection~\cite{girshick2014rich,ren2015faster}, semantic segmentation~\cite{long2015fully,ronneberger2015u}, and natural language processing~\cite{sutskever2014sequence,vaswani2017attention}. Many exemplary architectures have been proposed in image classification. For example, AlexNet~\cite{krizhevsky2012imagenet} and VGGNet~\cite{simonyan2014very} showed that the depth of convolutional neural networks is vital for achieving higher performance; ResNet~\cite{he2016deep} showed that identity-based skip connections are suitable for training deep neural networks; MobileNet~\cite{howard2017mobilenets,sandler2018mobilenetv2,howard2019searching} proposed the depthwise separable convolutions to build a lightweight model for edge devices. 

With the success of deep neural nets, the demand for deploying deep learning algorithms to the edge rises rapidly. Compared with cloud platforms, edge devices have the advantages of low cost, energy-saving, but with limited computation resources. Quantization~\cite{jacob2018quantization} is an essential technique to make the neural network run more efficiently on edge devices. Previous works~\cite{sandler2018mobilenetv2,iandola2016squeezenet,zhang2018shufflenet} address the issues and handcraft edged device-friendly models. However, since there are infinite candidate neural architectures, it is inefficient to find the optimal model relying on the manual trial and error method. As a result, neural architecture search (NAS) is proposed to find the optimal model architecture more efficiently using machine learning.

\begin{figure}[t]
\begin{center}
\includegraphics[width=1.0\linewidth]{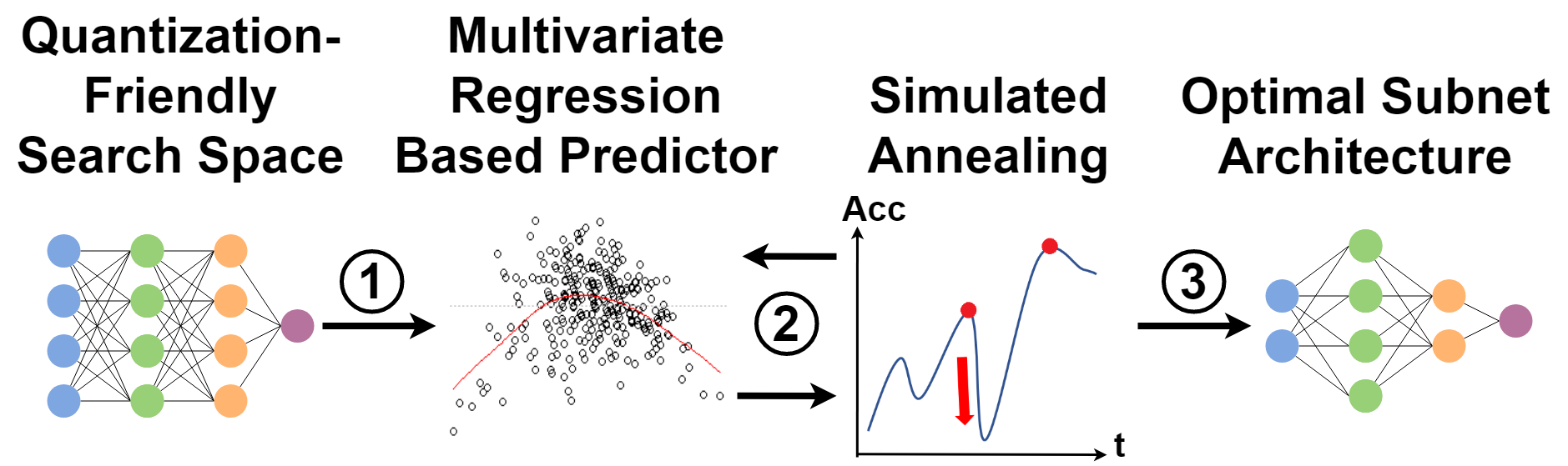}
\end{center}
\caption{We define a quantization-friendly search space (cf. Section~\ref{MethodB}), predict the subnet's performance by proposed multivariate regression (cf. Section~\ref{MethodC}), and use simulated annealing guided by our explainable predictors to avoid local optimal subnets and quickly find the global optimal subnet (cf. Section~\ref{MethodD}).
}
\label{img1}
\end{figure}

Neural architecture search is a technique for finding the optimal network architecture based on the search goal in the search space. The search goal can be accuracy, inference time, or any user-defined constraint. The most naïve method is to exhaustedly instance models with different architectures from the search space and then train each model to estimate its performance. Since there are countless permutations and combinations of architectures and it is unlikely to train every candidate model, methods based on reinforcement learning~\cite{zoph2016neural, zoph2018learning, tan2019mnasnet} were proposed to do NAS. More NAS methods that effectively reduce the requiring time for NAS were then proposed. For example, Progressive NAS~\cite{liu2018progressive} uses the sequential model-based optimization (SMBO) method as the search strategy; AmoebaNet~\cite{real2019regularized} proposed using the evolutionary algorithm to find the optimal network architecture. However, these NAS methods take a lot of GPU time to complete the training.

Recently, One-Shot NAS has been proposed to make NAS more efficient and effective. For example, ENAS~\cite{pham2018efficient} proposed to use the method of sharing weight in the training process of searching an architecture, so that the search time can be reduced to 16 GPU hours; DARTS~\cite{liu2018darts} proposed an algorithm of gradient based optimization for differentiable NAS; ProxylessNAS~\cite{cai2018proxylessnas} proposed an effective solution that can directly search the architectures for large-scale datasets and target hardware platforms. Additionally, Once-for-All~\cite{cai2019once} proposed a method that is different from the previous NAS. They decouple the training of the supernet from the architecture search and directly get a specialized subnet by selecting from the well-trained supernet without retraining. After the supernet training is completed, subnets are randomly sampled from the supernet to measure the performance, and then these data are used to do the architecture search. However, training the predictor required by the architecture search is time-consuming.

In this work, we propose a novel method for NAS named FOX-NAS, which has the advantages of being fast, on-device, and explainable. We continue the previous method~\cite{cai2019once} and reduce the time required for architecture search to complete the architecture search process directly on edge devices, as shown in Figure \ref {img1}.

The contribution of this work has four aspects:
\begin{itemize}
    \item [1)] 
    We adopt multivariate regression analysis as our predictors that reduce the time and data than the deep learning approach. In addition, the results of the parameters are explainable and controllable, which allows us to optimize the model for a variety of objectives, \eg, power consumption, accuracy, latency.
    \item [2)]
    We use simulated annealing as our search algorithm, which can complete the search within 1 minute and avoid suboptimal network structures.
    \item [3)]
    We design quantization-friendly search spaces that adapt for CPU and TPU so that the resulting models are easily deployed to edge devices.
    \item [4)] 
    Our extensive experiments are shown in Table~\ref{table1}, demonstrating that the architecture we found is 4.2\% higher than MobileNetV3-small~\cite{howard2019searching} under the same latency, and in the edge TPU environment, the accuracy of our model is also higher than MobileNetEdgeTPU.
\end{itemize}


\begin{table*}[t]
\begin{center}
\begin{tabular}{c|c|c|c|c|c}
\hline
Model                    & Search Strategy                                                         & Search Space                                                         & \begin{tabular}[c]{@{}c@{}}Performance \\ Estimation Strategy\end{tabular} & \begin{tabular}[c]{@{}c@{}}Training Cost \\ (GPU hours)\end{tabular} & \begin{tabular}[c]{@{}c@{}}Search Cost \\ (GPU hours)\end{tabular} \\ \hline
NASNet~\cite{zoph2018learning}          & \begin{tabular}[c]{@{}c@{}}Reinforcement\\ learning\end{tabular}        & arch                                                                 & Train and evaluate                & \multicolumn{2}{c}{48000N}                                                                                                               \\ \hline
MnasNet~\cite{tan2019mnasnet}         & \begin{tabular}[c]{@{}c@{}}Reinforcement\\ learning\end{tabular}        & arch                                                                 & Train and evaluate                & \multicolumn{2}{c}{40000N}                                                                                                               \\ \hline
AmoebaNet~\cite{real2019regularized}       & \begin{tabular}[c]{@{}c@{}}Evolution\\ algorithm\end{tabular}           & arch                                                                 & Train and evaluate                & \multicolumn{2}{c}{75600N}                                                                                                               \\ \hline
DARTS~\cite{liu2018darts}           & \begin{tabular}[c]{@{}c@{}}Gradient\\ optimization\end{tabular}         & arch                                                                 & Train and evaluate                & 250N                                                                 & 96N                                                                \\ \hline
ProxylessNAS~\cite{cai2018proxylessnas}    & \begin{tabular}[c]{@{}c@{}}Gradient\\ optimization\end{tabular}         & arch                                                                 & Train and evaluate                & 300N                                                                 & 200N                                                               \\ \hline
Once-for-All~\cite{cai2019once}    & \begin{tabular}[c]{@{}c@{}}Supernet/\\ Evolution algorithm\end{tabular} & arch                                                                 & Performance predictor             & 1200 + 40                                                            & $\approx$ 0                                                                 \\ \hline
FBNetV3~\cite{dai2020fbnetv3}         & \begin{tabular}[c]{@{}c@{}}Supernet/\\ NARS\end{tabular} & arch/recipe                                                          & Performance predictor             & 10700                                                                & $\approx$ 0                                                                 \\ \hline
\bf{FOX-NAS}                  & \begin{tabular}[c]{@{}c@{}}\bf{Supernet/}\\ \bf{Simulated annealing}\end{tabular} & \begin{tabular}[c]{@{}c@{}}\bf{Quantization}\\ \bf{friendly arch}\end{tabular} & \begin{tabular}[c]{@{}c@{}}\bf{Explainable}\\ \bf{performance predictor}\end{tabular} & \bf{1200 + 3.5}                                                           & \bf{$\approx$ 0}                                                                 \\ \hline
\end{tabular}
\end{center}
\caption{The search method comparison between FOX-NAS and the state-of-the-art NAS on ImageNet. We propose new methods in all three parts of NAS. FOX-NAS only needs to train the supernet once, which takes 1200 GPU hours, and then it only takes 3.5 hours to train performance predictors.}
\label{table1}
\end{table*}

\section{Preliminaries}

NAS can be divided into three parts: search space, search strategy, and performance estimation.

{\bf Search Space.\quad}Since there are infinite combinations of neural network architectures, we first need to define the architecture's scope, called search space. In image classification, the backbone of the convolution model architecture, such as ResNet~\cite{he2016deep}, MobileNet~\cite{sandler2018mobilenetv2, howard2019searching}, is used most frequently in NAS. In this work, we also adopt MobileNetV3~\cite{howard2019searching} as our backbone with quantization-friendly modules (\eg, change the activation function to ReLU6).

{\bf Search Strategy.\quad}The search strategy is a key to an optimal network search since the search space is enormous (approximately $4\times10^{22}$ in this work). In previous work, they used reinforcement learning~\cite{zoph2016neural, zoph2018learning, tan2019mnasnet}, SMBO~\cite{liu2018progressive}, and the evolution algorithm~\cite{real2019regularized}. In this work, to avoid the local optimal solutions, we use the simulated annealing algorithm as our search strategy. Compared with other search strategies, simulated annealing has a higher probability of finding the global optimal solution. Moreover, coupled with the guidance of multivariate regression analysis, we can find the optimal solution more quickly.

{\bf Performance Estimation.\quad}We evaluate the performance of the architectures sampled by the search strategy to find the optimal network architecture. However, this is infeasible, especially when the search space and target dataset are large. Therefore, proxy tasks (\eg, CIFAR-10) are often used to evaluate neural architecture performance. However, the optimal neural architectures found on the proxy tasks are not guaranteed to be the optimal architecture found in the target task~\cite{cai2018proxylessnas}. Once-for-all~\cite{cai2019once} proposed using a neural network to generate accuracy and latency predictors, and it took 40 GPU hours to collect data. In this work, we propose to use multivariate regression to generate predictors, which can collect enough data for training in just 3.5 GPU hours.

\begin{figure}[t]
\begin{center}
\includegraphics[width=1.0\linewidth]{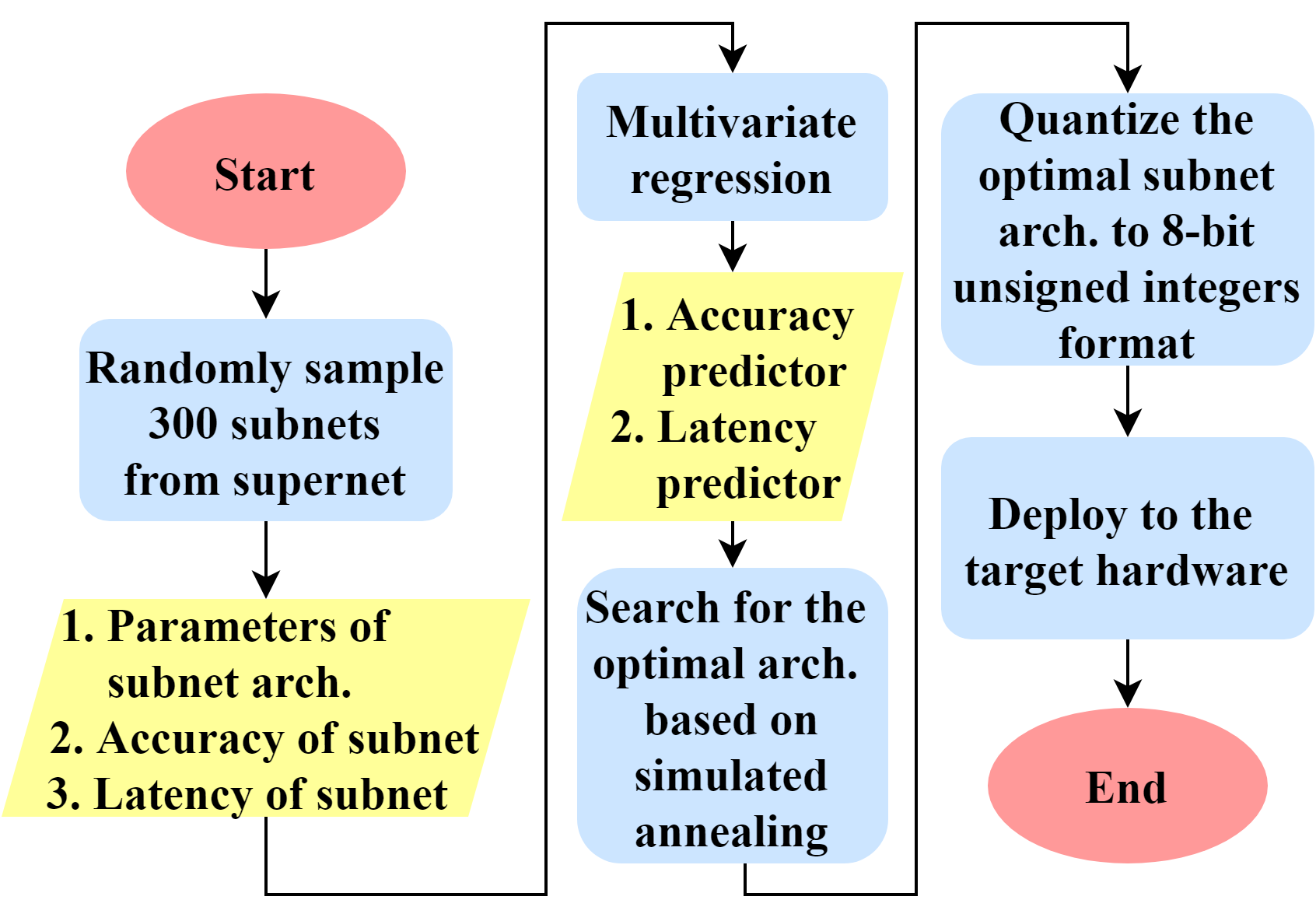}
\end{center}
\caption{Flowchart of FOX-NAS. We collect data of subnet information from our quantization-friendly supernet first, then use multivariate regression to generate performance predictors, and then use the simulated annealing algorithm to find the optimal neural network architecture. After quantized the searched subnet into 8-bit integer format, it can directly deploy the subnet to the specific edge device.}
\label{img2}
\end{figure}

\section{Method}

\subsection{Problem Statement}\label{Method}
Given a targeted latency on the specific hardware, we aim to find an optimal neural network, based on the neural architecture search (NAS) techniques, with the highest accuracy while meeting the constraints. Figure~\ref{img2} is our flow chart.

\begin{figure}
     \centering
     \begin{subfigure}[b]{0.38\linewidth}
         \centering
         \includegraphics[width=\linewidth]{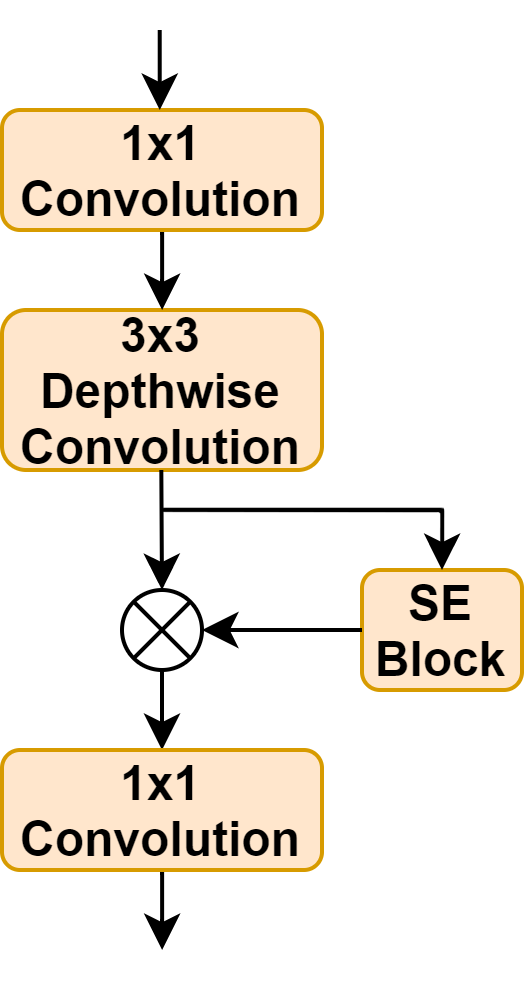}
         \caption{Type 1}
        \label{img3a}
     \end{subfigure}
     \hfill
     \begin{subfigure}[b]{0.23\linewidth}
         \centering
         \includegraphics[width=\linewidth]{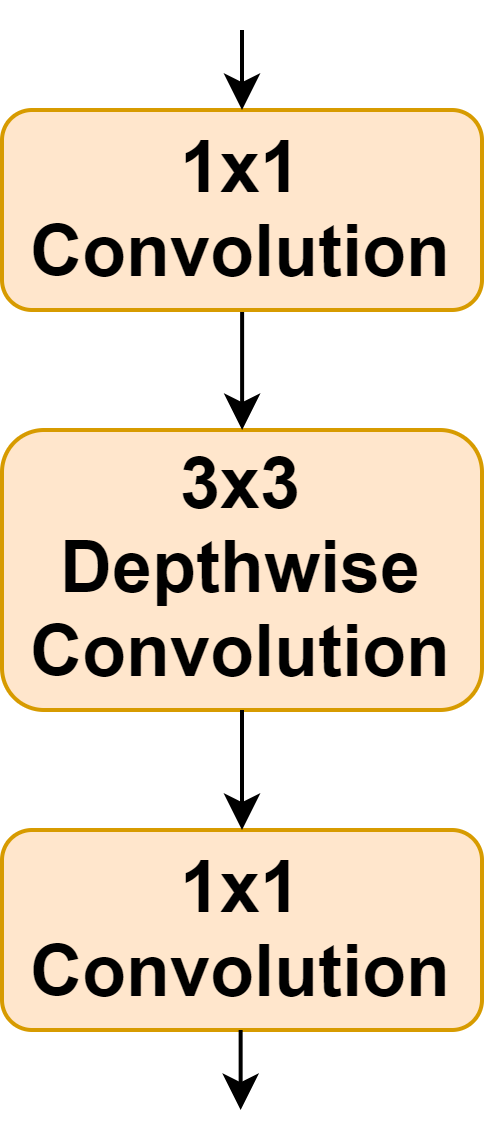}
         \caption{Type 2}
        \label{img3b}
     \end{subfigure}
     \hfill
     \begin{subfigure}[b]{0.23\linewidth}
         \centering
         \includegraphics[width=\linewidth]{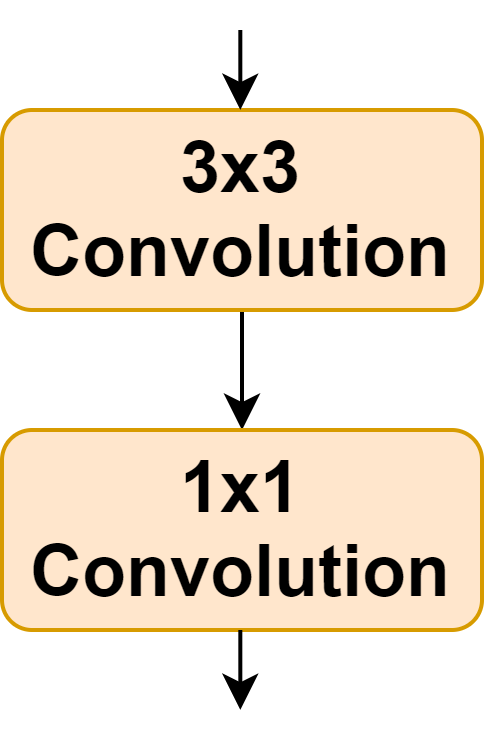}
         \caption{Type 3}
        \label{img3c}
     \end{subfigure}
        \caption{CNN units. FOX-NAS has three CNN units for different hardware.}
\label{img3}
\end{figure}

\subsection{Search Space}
\label{MethodB}
Regarding the famous CNN model architectures~\cite{sandler2018mobilenetv2, howard2019searching}, we also divide the CNN model into a sequence of units, and we have three types of CNN units, as shown in Figure~\ref{img3}. To make its operations faster and more efficient for different hardware, we design two different search spaces and replace the activation function with ReLU6 to make our neural architecture have the advantage of being quantization-friendly. We use the results of multivariate regression analysis, which allowed us to explain the impact of each control parameter on the performance of a neural network.


\begin{table}[]
\begin{center}
\begin{tabular}{c|c|c}
\hline
Search Space              & \begin{tabular}[c]{@{}c@{}}CPU\\ Backbone\end{tabular}  & \begin{tabular}[c]{@{}c@{}}TPU\\ Backbone\end{tabular} \\ \hline
Image Size Candidate      & 128$\sim$320 & 128$\sim$320 \\ \hline
Kernel Size Candidate     & 3, 5, 7      & 3            \\ \hline
Expansion Ratio Candidate & 2, 3, 4, 6   & 4, 6, 8      \\ \hline
Depth Candidate           & 2, 3, 4      & 3, 4, 5      \\ \hline
Type of CNN Unit          & 1, 2         & 2, 3         \\ \hline
\end{tabular}
\end{center}
\caption{FOX-NAS has two search spaces for different hardware. The adjustable architecture parameters of the search space include image size, kernel size, expansion ratio, depth, and type of CNN unit.}
\label{table2}
\end{table}

{\bf Search Space for CPU.\quad}The CPU-like hardware uses the first and second types of CNN units, as shown in Figure~\ref{img3a} and Figure~\ref{img3b}. The memory access and computation of CPU-like hardware are expensive, so the separable convolution is needed to reduce the computation of CNN. Therefore, we refer to the backbone of MobileNetV3~\cite{howard2019searching} and the supernet of Once-for-All~\cite{cai2019once}. Table~\ref{table2} lists the candidate of all the architecture parameters in our search space. We found that the number of channels of the previous units significantly impacted the latency on the CPU-like hardware through the multivariate regression analysis. As a result, when designing the search space, we set the minimum expansion ratio to 2 while making more choices for our subnet. In addition, to find the subnet with a broader accuracy range, our input image size is changed from 128 $\times$ 128 to 320 $\times$ 320. Finally, we adopt the quantization-friendly activation function to achieve a better quantization effect, which enables our subnet to control the accuracy loss within 1\% when quantizing. The number of different neural network architectures in our CPU search space is approximately $4\times10^{22}$.

{\bf Search Space for TPU.\quad}The TPU-like hardware uses the second and third types of CNN units, as shown in Figure~\ref{img3b} and Figure~\ref{img3c}. The TPU-like hardware has a high degree of parallelism in matrix operations, so the higher the processor utilization, the higher the computing efficiency. Therefore, we refer to the model architecture of MobilenetEdgeTPU and replace the separable convolution of the previous unit with the traditional convolution. As shown in Table~\ref{table2}, we fix the kernel size of search space at 3 $\times$ 3 because the 5 $\times$ 5 and 7 $\times$ 7 kernel sizes are not friendly for the edge TPU. TPU has excellent parallelism, so we change the candidate of expanding ratio to 4, 6, 8. The number of different neural network architectures in our TPU search space is approximately $3\times10^{13}$. The effect of different backbones on different hardware is different, and the related experimental results and analysis are in Section~\ref{ExperimentsA}.

\subsection{Performance Prediction Based on Multivariate Regression}
\label{MethodC}
Multivariate regression analysis is a machine learning algorithm based on supervised learning, which can analyze multiple variables (\eg, the impact of each layer's kernel size and expansion ratio on performance). We adopt multivariate regression to generate performance predictors. This method is not data-hungry, and it is fast and explainable. Compared with the method based on deep learning, multivariate regression analysis can better understand the effect of each variable on the results and create promising predictors with much fewer data. Furthermore, the results of multivariate regression analysis make the parameters of a network architecture explainable, which provides a hint of twisting the architecture for target constraints. We trained a total of 7 predictors for different image sizes, and each predictor only needs 300 pieces of data to train. Using a consumer-level GPU only takes 3.5 GPU hours (the previous method needs to collect 16K data, a total of 40 GPU hours~\cite{cai2019once}). To collect the training data for multivariate regression, we first randomly sample different sub-networks from the super-network and record their network architectures, characterized by 25 variables, including the number of layers, widths, and kernel sizes. We then use 50K validation data to measure the accuracy of each sub-network. Meanwhile, we run each sub-network on the target hardware to collect the latency data.

Assuming that the estimated multivariate linear regression model is:

\begin{dmath}
\widehat{Y} = \widehat{\beta}_0 + \widehat{\beta}_1X_1 + \widehat{\beta}_2X_2 + \text{…} + \widehat{\beta}_kX_k
\end{dmath}

$\beta_i$ are slope parameters or called correlation coefficients, representing the impact of the variable, $Y$ is the dependent variable, $X_i$ are independent variables, the symbol $\wedge$ indicates an estimate for the variables. Moreover, taking our CPU search space as an example, the performance prediction model equation can be expressed as follows:

\begin{dmath}
\widehat{Perf.} = \widehat{\beta}_0 + \widehat{\beta}_{D_1}D_1 + \widehat{\beta}_{E^{avg}_1}E^{avg}_1 + \widehat{\beta}_{K^{avg}_1}K^{avg}_1 +
\\\widehat{\beta}_{E_{1,2}}E_{1,2} + \widehat{\beta}_{K_{1,2}}K_{1,2} + \widehat{\beta}_{E^{total}_1}E^{total}_1 + 
\\\widehat{\beta}_{E^{total}_1{\cdot}D_1}E^{total}_1 \cdot D_1 + \widehat{\beta}_{K^{avg}_1{\cdot}D_1}K^{avg}_1 \cdot D_1 + \text{…} + 
\\\widehat{\beta}_{K^{avg}_m{\cdot}D_m}K^{avg}_m \cdot D_m
\end{dmath}

FOX-NAS-CPU has five types of CNN units with different input channels, so $m$ equals 5 in this case. $j$ represents the type of CNN units, $D_j$ are the number of depth of each CNN unit, $E^{avg}_j$ are the average expansion ratio of each CNN unit, $K^{avg}_j$ are the average kernel size of each CNN unit, $E_{j,j+1}$ and $K_{j,j+1}$ are the expand ratio and kernel size of each CNN unit with different input and output channels, $E^{total}_j$ are the total number of expansion ratio of each CNN unit, and $E^{total}_j \cdot D_j$ and $K^{avg}_j \cdot D_j$ represent the interaction between total expansion ratio and number of depth of each CNN unit and the interaction between the average kernel size and number of depth of each CNN unit, respectively.

In addition, we can analyze and explain our neural network architecture through some statistics. $S_{{\widehat{{\beta}i}}}$ are the standard deviation of estimate coefficients $\widehat{{\beta}i}$, called the standard error. The t-value is a commonly used statistic, and its formula is as follows:

\begin{dmath}
t = \frac{\widehat{\beta}_i}{S_{{\widehat{\beta}_i}}}
\end{dmath}

The p-value is the probability density value $\geq$ t-value under the T-distribution, representing whether the impact of the variable on the output variable is highly correlated. For example, the p-value less than 0.05 indicates that the variable is highly correlated with the output variable. In addition, we can look up the t-value in the T-distribution table with the given degrees of freedom to get the p-value. The $R^{2}$ value measures the percentage of the variation in $Y$ being explained by the fitted regression model. Thus, the larger the $R^{2}$ value, the better the fit of the regression model, and its formula is as follows:

\begin{dmath}
TSS = \sum_{i=1}^{n} (Y_i - \overline{Y})^{2}
\end{dmath}

\begin{dmath}
SSE = \sum_{i=1}^{n} (Y_i - \widehat{Y}_i)^{2}
\end{dmath}

\begin{dmath}
R^{2} = 1 - \frac{SSE}{TSS}
\end{dmath}

$TTS$ is the total sum of squares, $SSE$ is sum of squares errors, $\overline{Y}$ is the mean of $Y$. When adding more independent variables, $R^{2}$ will be larger, showing an overestimation phenomenon. Adjusted $R^{2}$ has been adjusted in degrees of freedom to avoid the expansion, and its formula is as follows:

\begin{dmath}
R_{adj}^{2} = 1 - \frac{SSE \times (n - 1)}{TSS \times (n - k)}
\end{dmath}

$n$ is the number of collected data, $k$ is the number of correlation coefficients. The adjusted $R^{2}$ value of the regression model we generate can reach above 92, indicating that our predictor is accurate. The variables in the predictor are highly correlated and explainable, so we can more effectively control performance by adjusting the architecture parameters.

When an optimal subnet is proposed from our predictors, we can further twist the architecture to make the performance of the subnet meet our target constraint since the impact of each architecture parameter on the subnet performance is explainable. We choose the architecture parameter with a sufficiently small p-value and then compare the correlation coefficients of the parameter to make a precise and efficient adjustment based on the target constraint. 

For example, suppose that the latency constraint we set is 60 ms, but the subnet we found through the predictor is 60.3 ms, which exceeds the constraint of 0.3 ms. We need to adjust the parameters of this subnet architecture manually. Suppose we find the p-value of $E^{avg}_1$ is close to 0 in both the accuracy and latency predictors, which means that the impact of $E^{avg}_1$ on the performance is highly correlated. Moreover, suppose the correlation coefficients $\widehat{\beta}$ of $E^{avg}_1$ is very small in the accuracy predictor, which means that $E^{avg}_1$ only has little impact on accuracy. Suppose the $\widehat{\beta}$ of $E^{avg}_1$ is 0.4 in the latency predictor, which means that the latency can be reduced by 0.4 ms when $E^{avg}_1$ is reduced by 1. As a result, we can manually twist $E^{avg}_1$ to achieve the target latency.

\begin{figure}
     \centering
     \begin{subfigure}[b]{0.49\linewidth}
         \centering
         \includegraphics[width=\linewidth]{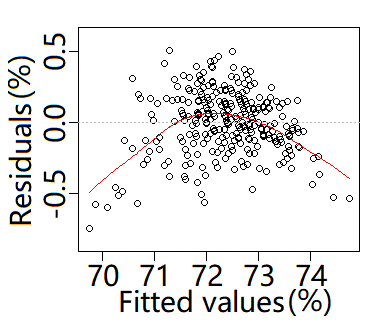}
         \caption{Residual plot}
        \label{img10}
     \end{subfigure}
     \hfill
     \begin{subfigure}[b]{0.49\linewidth}
         \centering
         \includegraphics[width=\linewidth]{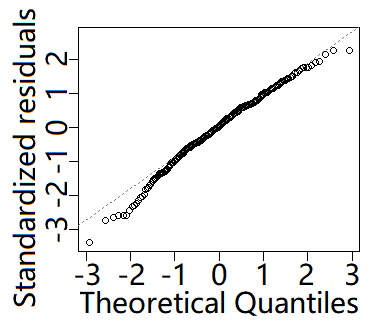}
         \caption{Normal Q-Q plot}
        \label{img11}
     \end{subfigure}
        \caption{{\bf(a)} The red line in the residual plot is our model, and the points in the figure are the collected data. The fitted value is between 70 and 74, and the maximum residual is only 0.5, which is a small residual, less than 1\% of the fitted value. {\bf(b)} There is a 45° straight dashed line in the normal Q-Q plot. If the data is normally distributed, the point will fall on the 45° reference line. The bottom end of the Q-Q plot deviates from the straight line, but the upper end does not deviate, then we can say that it is a left-skewed distribution.}
\label{img10_11}
\end{figure}

Figure~\ref{img10} shows the residual plot of our accuracy predictor. A fitted value $\widehat{Y}$ is the model's prediction of the response value when we input the values of the predictor. The residuals $e$ are equal to the difference between the ground truth and the fitted value:

\begin{dmath}
e = Y - \widehat{Y}
\end{dmath}

The normal quantile-quantile plot is used to evaluate whether our residuals are normally distributed. Figure~\ref{img11} shows our normal Q-Q plot. These analysis graphs represent the high reliability of our model.

\subsection{Search Strategy Based on Simulated Annealing}
\label{MethodD}
\begin{figure}[t]
\begin{center}
\includegraphics[width=1.0\linewidth]{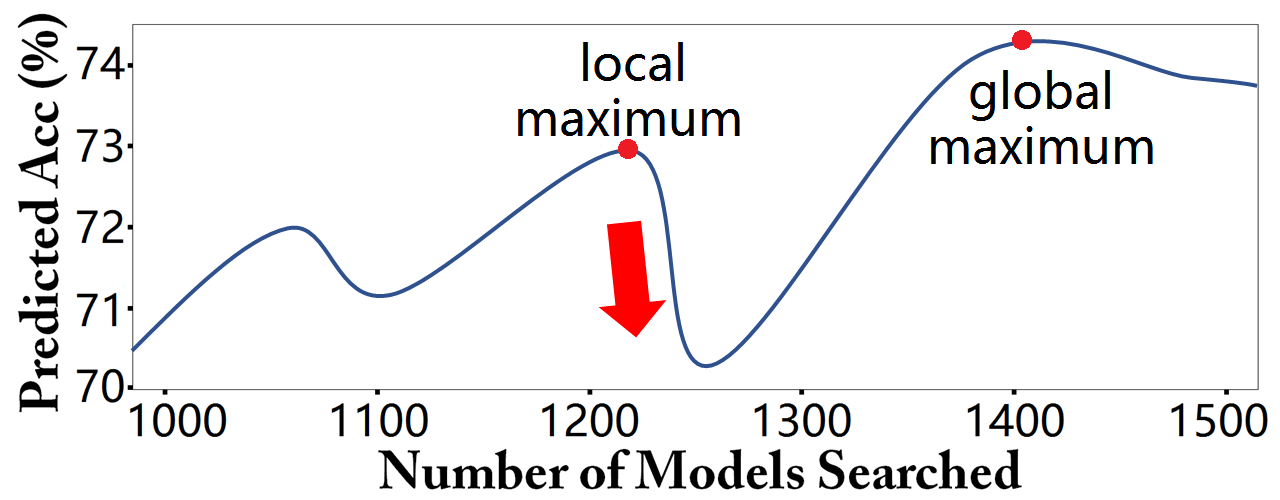}
\end{center}
\caption{When we find the local maximum, if the local maximum is far away from the global maximum, we need to go through a period of inferior results before we can find the global maximum solution. The simulated annealing algorithm accepts unsatisfactory results, so it has a higher probability of finding the global maximum than the local maximum compared to other algorithms.}
\label{img14}
\end{figure}

Simulated annealing is an algorithm based on probability to find the optimal solution under the objective function. It is usually used when the search space is discrete. Because it can accept unsatisfactory results during the search process, it can find the global optimal solution rather than the local optimal solution compared to other algorithms, as shown in Figure~\ref{img14}.

We use simulated annealing as our search strategy. As shown in Figure~\ref{img2}, we can find the optimal neural architecture in one minute after using multivariate regression to generate the performance predictors, combined with the simulated annealing algorithm. The simulated annealing method we used is summarized in Algorithm~\ref{a1}. Under the constraint of computational resources, we use the guidance of multivariate regression analysis to the sample neural architecture, aiming to find the neural architecture with the best performance. In the process of searching based on simulated annealing, we accept the neural architectures with inferior performance, but as the number of architectures searched increases, we reduce the acceptance of unsatisfactory architectures until the algorithm converges. This is also a feature of simulated annealing, which has the advantage of a higher probability of finding the global optimal solution. 

With the guidance of multivariate regression analysis, we can find the optimal model more quickly. We give each architecture parameter a weight, representing the probability of selecting this parameter when sampling a subnet architecture. In the original method, the weight of each parameter is 1, which means that the probability of selecting each parameter is equal. However, through multivariate regression analysis, we know the impact of each architecture parameter on the performance of the subnet. Accordingly, in the search process, we have two sets of weights. In the early stage of the search, we use the first set of weights, which are very large for a few architecture parameters. The purpose is to find a preliminary solution first. Then we change to the second set of weights, which are almost the same for all architecture parameters, to find the global optimal solution. The related experimental results and analysis are in Section~\ref{ExperimentsB}.

\begin{algorithm}[]
\SetKwInput{KwInput}{Input}                
\SetKwInput{KwOutput}{Output}              
\DontPrintSemicolon

  \KwInput{latency constraint $L$}
  \KwOutput{optimal model $M$}

  \SetKwFunction{FMain}{Main}
  \SetKwFunction{FSampleModel}{SampleModel}

  // When sampling the model architecture, we select $T$ architecture parameters to make changes.\;
  // We have two sets of weights representing the probability of selecting each architecture parameter when sampling the model architecture.\;
  // We have a probability formula to evaluate whether to accept the architecture with inferior estimation performance.\;\;
  \SetKwProg{Fn}{Function}{:}{}
  \Fn{\FSampleModel{$M$, $T$, $weight$}}{
        $tmp.lat$ $\leftarrow$ $\infty$\;
        \While{$tmp.lat$ $\geq$ $L$}{
            $tmp.arch$$\leftarrow$SampleArch($M, $T$, weight$)\;
            $tmp.acc$ $\leftarrow$ AccPredictor($tmp.arch$)\;
            $tmp.lat$ $\leftarrow$ LatPredictor($tmp.arch$)\;
        }
        \KwRet $tmp$\;
  }
  \;

  \SetKwProg{Fn}{Function}{:}{\KwRet}
  \Fn{\FMain}{
        Get an initial model $M$\;
        Get an initial temperature $T$ $\textgreater$ $0$\;
        Get a counter $C$ = $0$\;
        Get the constants $k$, $n$\;
        $weight$ $\leftarrow$ weights of the early stage (guided by our explainable predictors)\;
        \While{$T$ $\textgreater$ $0$}{
            $C$ $\leftarrow$ $C$ + 1\;
            $r$ $\leftarrow$ Random(0,1)\;
            $M\_new$ $\leftarrow$ SampleModel($M$, $T$, $weight$)\;
            $\triangle$ $\leftarrow$ $M\_new.acc$\ \ -\ \ $M.acc$\;
            \uIf{$\triangle$ $\textgreater$ $0$ \bf{or} $r$ $\textgreater$ exp(-$\triangle$ / $kT$)}{
                $M$ $\leftarrow$ $M\_new$\;
            }
            \Else{
                \If{C \% $n$ == 0}{
                    Decrease the temperature $T$\;
                }
            }
            \If{$C$ == $2n$}{
                $weight$ $\leftarrow$ new set of weights searched\;
            }
        }
        \KwRet $M$\;
  }
\caption{NAS based on Simulated Annealing}
\label{a1}
\end{algorithm}

\begin{figure}[t]
\begin{center}
\includegraphics[width=1.0\linewidth]{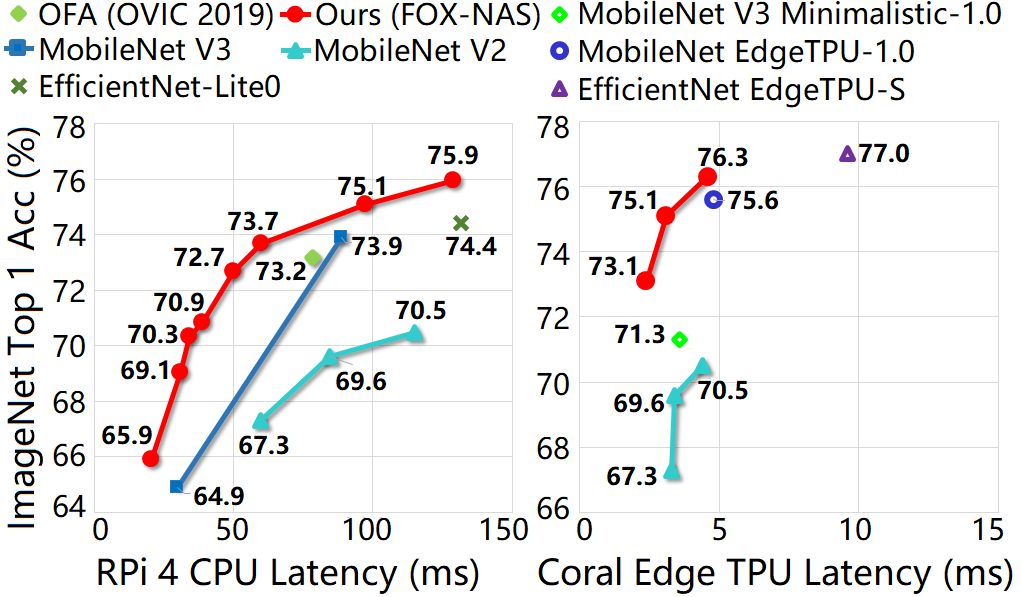}
\end{center}
\caption{FOX-NAS achieves SOTA performance on ARM CPU and edge TPU.}
\label{img4}
\end{figure}

\begin{table}[]
\setlength{\tabcolsep}{4.8pt}
\begin{center}
\begin{tabular}{ccccc}
\hline
Model                      & \begin{tabular}[c]{@{}c@{}}Top-1\\ (\%)\end{tabular} & \begin{tabular}[c]{@{}c@{}}GPU\\ latency\\ (ms)\end{tabular} & \begin{tabular}[c]{@{}c@{}}Image\\ size\end{tabular} & \begin{tabular}[c]{@{}c@{}}Params\\ (M)\end{tabular} \\ \hline
MNASNet-0.5~\cite{tan2019mnasnet}       & 68.9                                                 & 13                                                           & 224        & 2.1                                                      \\
MNASNet-0.75~\cite{tan2019mnasnet}      & 73.3                                                 & 20                                                           & 224        & 2.9                                                      \\
MNASNet-1.0~\cite{tan2019mnasnet}       & 75.2                                                 & 24                                                           & 224        & 3.9                                                      \\ \hline
MobileNetV2~\cite{sandler2018mobilenetv2}       & 72                                                   & 40                                                           & 224        & 3.5                                                      \\ \hline
MobileNetV3-S~\cite{howard2019searching} & 67.5                                                 & 10                                                           & 224        & 2.9                                                      \\
MobileNetV3-L~\cite{howard2019searching} & 75.2                                                 & 25                                                           & 224        & 5.4                                                      \\ \hline
EfficientNet-B0~\cite{tan2019efficientnet}   & 76.1                                                 & 47                                                           & 224        & 5.3                                                      \\ \hline
FBNetV2-F3~\cite{wan2020fbnetv2}        & 73.2                                                 & 18                                                           & 224        & 6.9                                                      \\
FBNetV2-F4~\cite{wan2020fbnetv2}        & 76                                                   & 25                                                           & 224        & 7.0                                                      \\
FBNetV2-L1~\cite{wan2020fbnetv2}        & 77.2                                                 & 31                                                           & 224        & 8.5                                                      \\ \hline
FBNetV3-A~\cite{dai2020fbnetv3}         & 79.1                                                 & 33                                                           & 224        & 8.6                                                      \\ \hline
FairNAS-A~\cite{chu2019fairnas}         & 75.3                                                 & 28                                                           & 224        & 4.6                                                      \\
FairNAS-A-SE~\cite{chu2019fairnas}      & 77.5                                                 & 34                                                           & 224        & 5.9                                                      \\ \hline
ProxylessNAS~\cite{cai2018proxylessnas}      & 75.1                                                 & 25                                                           & 224        & 5.1                                                      \\ \hline
OFA-1080Ti-15~\cite{cai2019once}   & 73.8                                                 & 13                                                           & 144        & 6.0                                                      \\
OFA-1080Ti-22~\cite{cai2019once}   & 75.3                                                 & 17                                                           & 188        & 5.2                                                      \\
OFA-1080Ti-27~\cite{cai2019once}   & 76.4                                                 & 22                                                           & 188        & 5.2                                                      \\ \hline
\bf{FOX-NAS-TPU-A}              & 73.9                                                 & 12                                                           & 192        & 4.1                                                      \\
\bf{FOX-NAS-TPU-B}              & 75.3                                                 & 17                                                           & 192        & 5.3                                                      \\
\bf{FOX-NAS-TPU-C}              & 76.3                                                 & 22                                                           & 224        & 5.3                                                      \\ \hline
\end{tabular}
\end{center}
\caption{Comparison with popular models on Nvidia 2080Ti GPU. On the Nvidia 2080Ti GPU, FOX-NAS can achieve the best accuracy under the three different latency constraints.}
\label{table3}
\end{table}

\begin{figure}[t]
\begin{center}
\includegraphics[width=1.0\linewidth]{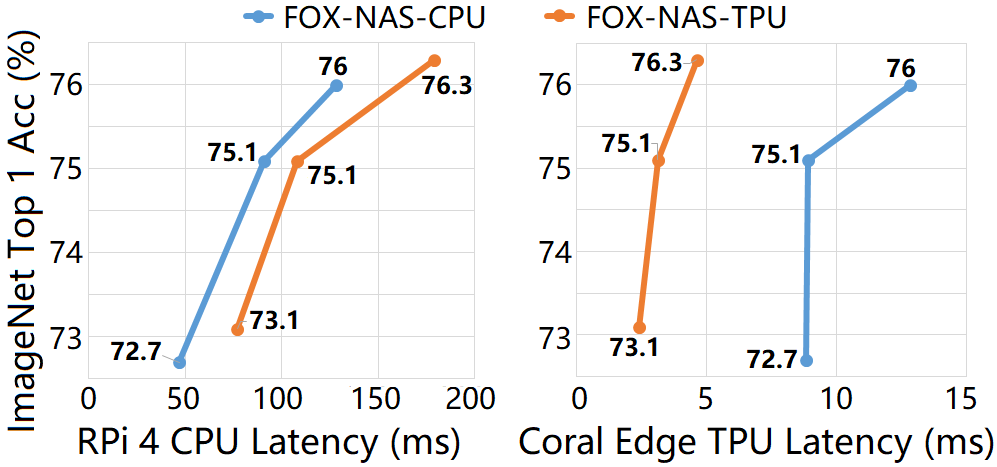}    
\end{center}
\caption{Latency comparison between CPU backbone and TPU backbone on different hardware. The backbone of the CPU runs more efficiently on the CPU than the backbone of the TPU. On the contrary, the backbone of the TPU runs more efficiently on edge TPU.}
\label{img7}
\end{figure}

\begin{figure}[t]
\begin{center}
\includegraphics[width=1.0\linewidth]{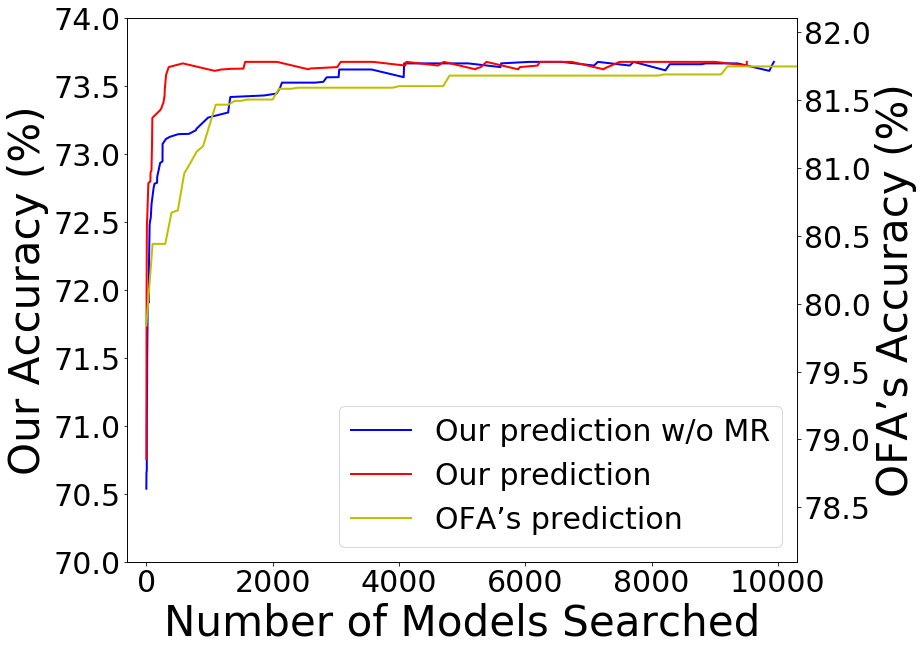}
\end{center}
\caption{Performance comparison of search strategy between OFA and FOX-NAS. FOX-NAS uses simulated annealing and multivariate regression analysis to make our search process converge faster. The label on the left y-axis is the prediction of our accuracy predictor, and the label on the right is the prediction of OFA's predictor. OFA's predictors are provided by their GitHub. Because OFA and we have different hardware constraints, there will be a gap between the predicted results. To facilitate comparison, we shift the predictions.}
\label{img12}
\end{figure}

\section{Experiments}
In this section, we compare FOX-NAS on different hardware with some popular neural networks. In addition, we compare the performance of FOX-NAS with different backbones on different hardware. Finally, there is a performance comparison between different search methods. Our experiment is performed on image classification using ImageNet~\cite{deng2009imagenet}.

\subsection{Comparison of Model Performance with Different Hardware and Constraints}
\label{ExperimentsA}
We compare the performance of the models on different hardware, including cloud GPU, edge TPU, and edge CPU. In the experiment on GPU, the latency is measured with batch size 64 and 32-bit floating-point format on Nvidia 2080Ti with Pytorch 1.8.1 + CUDA 11.0. As for edge CPU and edge TPU, we use the quantization tool of Tensorflow 2, and the latency is measured with batch size 1 and 8-bit unsigned integer format on the Raspberry pi 4 ARM Cortex-A72 CPU and Coral USB TPU accelerator.

{\bf Latency Comparison on Edge CPU.\quad}Figure~\ref{img4} shows the performance comparison between FOX-NAS and other famous and open-source integer neural network models. The hardware used for the measurement is the ARM CPU on the Raspberry Pi 4. FOX-NAS can generate subnets with an extensive accuracy range, from 76\% to 66\%, and each subnet performs well. Under the same accuracy level, FOX-NAS can be 240\% faster than MobileNetV2~\cite{sandler2018mobilenetv2}, 50\% faster than MobileNetV3~\cite{howard2019searching}, and 40\% faster than EfficientNet-Lite0~\cite{tan2019efficientnet}. Under the same latency level, FOX-NAS can be 6.4\% more accurate than MobileNetV2, 4.2\% more accurate than MobileNetV3, and 1.5\% more accurate than EfficientNet-Lite0.

{\bf Latency Comparison on Different Hardware.\quad}Unlike the hardware architecture of edge CPU, architecture such as edge TPU and cloud GPU has powerful parallel computing capabilities. Therefore, we propose different search spaces for different hardware to make the neural network operation more efficient. Figure~\ref{img4} shows the performance comparison between FOX-NAS and other popular models on edge TPU. The Edge TPU Compiler version used in our experiment is 2.0.291256449. Under the same latency level, FOX-NAS is 5.8\% more accurate than MobileNetV2~\cite{sandler2018mobilenetv2} and performs better than MobileNetEdgeTPU. Because the cache on the Coral USB TPU accelerator is small, only about 6 MB, and models over 6 MB will cause much damage to the latency, so we did not search for a larger model to evaluate.

Table~\ref{table3} show the performance comparison of FOX-NAS and other models on GPU. Because the edge TPU has some restrictions on neural network models, many models cannot be directly deployed on this device. To compare with more models, we deployed FOX-NAS with TPU based search space on the GPU. Compared with the Once-for-All~\cite{cai2019once} models, FOX-NAS can achieve almost the same performance while reducing the training cost by 36.5 GPU hours. Under the same accuracy level, FOX-NAS can be 47\% faster than MobileNetV3~\cite{howard2019searching} and 50\% faster than FBNetV2-F3~\cite{wan2020fbnetv2}.

{\bf Latency Comparison of Different Search Space on Different Hardware.\quad}It is mentioned in Section~\ref{MethodB} that different hardware requires different neural network architecture designs to be more efficient. Therefore, we propose two different search spaces: CPU-based and TPU-based. Figure~\ref{img7} is a comparison diagram of the latency results of running the two backbone subnets on the edge CPU and the edge TPU, respectively. The backbone of the CPU runs more efficiently on the CPU than the backbone of the TPU. On the contrary, the backbone of the TPU runs more efficiently on edge TPU. The reason is that the computing power on the CPU is the bottleneck, and the edge TPU has strong computing power, so the memory access is the bottleneck of the edge TPU.

\subsection{Performance Comparison of Different Search Methods}
\label{ExperimentsB}
We compare the search methods between FOX-NAS and Once-for-All~\cite{cai2019once}. In addition, we compare the performance of search using simulated annealing algorithm with or without multivariate regression analysis guidance.

{\bf Comparison of the Search Methods Between FOX-NAS and Once-for-All (OFA).\quad}Figure~\ref{img12} shows the comparison between our search and OFA search method. We use multivariate regression to generate performance predictors and then use simulated annealing coupled with guidance from multivariate regression analysis as our search strategy. OFA used predictors based on the neural network and used the evolutionary algorithm as the search strategy. In this experiment, the target hardware we searched for was the ARM CPU on the Raspberry Pi 4, and OFA was the CPU on the Samsung Note 10 mobile phone. We recorded the performance predicted by FOX-NAS and OFA during the search process. OFA requires 50,000 points to finish the search, so we only selected the first 10,000 points for comparison with FOX-NAS for easy comparison. As shown in Figure~\ref{img12}, FOX-NAS can search for the global optimal solution more quickly. Because the simulated annealing algorithm has the property of accepting inferior search results during the search process, it has more oscillations than the evolutionary algorithm during the convergence process.

Figure~\ref{img12} also compares the search performance of simulated annealing with or without multivariate regression analysis. After using the guidance of multivariate regression analysis, the simulated annealing can quickly find the global optimal solution near only a few points. In contrast, the simulated annealing search requires more points to find the global optimal solution without multivariate regression analysis.

\section{Conclusion}

We propose FOX-NAS, a novel neural architecture search method with fast, on-device, and explainable advantages. Unlike the previous approach, we reduced the time required to generate the performance predictors to complete in 3.5 GPU hours. Moreover, we reduced the movement steps required in the search process. Experiments on different hardware showed that the model obtained by our approach has good performance. Thus, we expect our work can reduce the cost of neural architecture search.

{\small
\bibliographystyle{ieee_fullname}
\bibliography{final}
}

\end{document}